\documentclass[conference, 9pt]{IEEEtran}
\IEEEoverridecommandlockouts
\usepackage{cite}
\usepackage{hyperref}
\usepackage{url}
\usepackage{makecell}
\usepackage{multirow}
\usepackage{xcolor}
\usepackage{relsize}
\usepackage{amsmath,amssymb,amsfonts}
\usepackage{algorithmic}
\usepackage{graphicx}
\usepackage{textcomp}
\usepackage{xcolor}

\ifCLASSOPTIONcompsoc
\usepackage[caption=false,font=normalsize,labelfon
t=sf,textfont=sf]{subfig}
\else
\usepackage[caption=false,font=footnotesize]{subfig}
\fi

\usepackage{booktabs}
\def\BibTeX{{\rm B\kern-.05em{\sc i\kern-.025em b}\kern-.08em
    T\kern-.1667em\lower.7ex\hbox{E}\kern-.125emX}}

\newcommand{\usermark}{$\mathtt{<USER>}$}%
\newcommand{\asstmark}{$\mathtt{<ASST>}$}%

\begin{document}

\title{AudioLens: A Closer Look at Auditory Attribute Perception of Large Audio-Language Models}

\author{\IEEEauthorblockN{Chih-Kai Yang}
\IEEEauthorblockA{\textit{National Taiwan University} \\
Taipei, Taiwan \\
chihkaiyang1124@gmail.com}
\and
\IEEEauthorblockN{Neo Ho}
\IEEEauthorblockA{\textit{National Taiwan University} \\
Taipei, Taiwan \\
neo1221ho@gmail.com}
\and
\IEEEauthorblockN{Yi-Jyun Lee}
\IEEEauthorblockA{\textit{National Taiwan University} \\
Taipei, Taiwan \\
james52231@gmail.com}
\and
\IEEEauthorblockN{Hung-yi Lee}
\IEEEauthorblockA{\textit{National Taiwan University} \\
Taipei, Taiwan \\
hungyilee@ntu.edu.tw}
}

\maketitle

\begin{abstract}
Understanding the internal mechanisms of large audio-language models (LALMs) is crucial for interpreting their behavior and improving performance. This work presents the first in-depth analysis of how LALMs internally perceive and recognize auditory attributes. By applying vocabulary projection on three state-of-the-art LALMs, we track how attribute information evolves across layers and token positions. We find that attribute information generally decreases with layer depth when recognition fails, and that resolving attributes at earlier layers correlates with better accuracy. Moreover, LALMs heavily rely on querying auditory inputs for predicting attributes instead of aggregating necessary information in hidden states at attribute-mentioning positions. Based on our findings, we demonstrate a method to enhance LALMs. Our results offer insights into auditory attribute processing, paving the way for future improvements.
\end{abstract}



\section{Introduction}

Recent advances in large language models (LLMs)~\cite{zhao2023survey, grattafiori2024llama, hurst2024gpt} have rapidly extended into the auditory domain, leading to large audio-language models (LALMs)~\cite{gong2024listen, qwen, qwen2, desta2, ghosh2024gama, desta, speechcopilot, gong2023joint, tangsalmonn, wang2024blsp, hu2024wavllm} that integrate auditory and textual understanding. These models support a broad spectrum of tasks, ranging from fundamental auditory perception, such as emotion recognition and language identification, to complex reasoning and interactive dialogue. As a result, extensive benchmarks have been established to comprehensively evaluate their capabilities~\cite{dynamicsuperb2, airbench, mmau, sdeval, wang2024audiobench, sakura, yang2025towards}.

While task-level evaluations offer useful insights~\cite{dynamicsuperb2, mmau, lin2025preliminary}, understanding the internal mechanisms of models is increasingly important. In LLM research, interpretability studies have elucidated how linguistic knowledge~\cite{tang2024language, zhao2023unveiling}, reasoning processes~\cite{yang2024large, biran2024hopping, yu2025back}, and world knowledge~\cite{dai2022knowledge, yu2024neuron} are internally represented, guiding model improvements. However, knowledge of how LALMs process auditory information remains limited. Existing studies focus on LALMs' high-level behaviors like biases~\cite{lin2024listen} or hallucinations~\cite{leng2024curse, kuan2025can}, without studying internal representations or processing dynamics.

To bridge this gap, we present the first study of auditory information processing in LALMs, focusing on auditory attribute perception, which is essential for many applications. Auditory attributes refer to properties of a sound, such as the speaker’s gender, emotional state, spoken language, or the type of animal producing the sound. Using the Logit Lens technique~\cite{logit_lens}, a training-free vocabulary projection method~\cite{geva2022transformer, din2024jump, tunedlens} effective for interpreting LLMs and multimodal models, we analyze how these attributes are encoded and resolved across layers and token positions in three state-of-the-art LALMs.

We find that attribute information does not steadily increase with layer depth; instead, it often drops sharply at certain layers before recovering later. This reflects two opposing dynamics: for correctly recognized samples, information rises with depth; for difficult ones, it peaks midway but diminishes in deeper layers, causing prediction errors. Furthermore, there is a generally negative correlation between the layer at which attribute information is resolved and prediction accuracy, indicating that when models resolve attribute information at earlier layers, more subsequent layers are available to refine this information, which leads to higher prediction accuracy.

We also compare information across token positions, finding that though attributes are previously mentioned, information aggregated at these preceding positions is insufficient for accurate prediction. Instead, LALMs heavily rely on querying auditory inputs directly. This result explains why LALMs struggle with complex reasoning tasks~\cite{sakura}. Based on our findings, we propose to enrich deeper-layer representations with earlier attribute-rich representations, boosting prediction accuracy with a 16.3\% relative improvement without training.

Our contributions are: (1) the first study of internal information processing in LALMs; (2) revealing layer-wise information dynamics and their relation to recognition accuracy; (3) analyzing information flow across tokens to identify the information sources for attribute predictions; and (4) introducing a novel improvement method based on these findings. Our work advances understanding of LALMs’ internal mechanisms and suggests directions for future enhancement. Code will be available at \url{https://github.com/ckyang1124/AudioLens}.

\section{Related Works}
\subsection{Understanding Auditory Foundation Models}

Before LALMs emerged, many studies analyzed auditory foundation models beyond task-level evaluation~\cite{superb, tsai2022superb, turian2022hear, yuan2023marble, zerospeech, cszs, csasrst, mlsuperb}. For self-supervised learning (SSL) models~\cite{hubert, chang2022distilhubert, schneider2019wav2vec, wav2vec2, xlsr}, several studies have performed layer-wise~\cite{pasad2021layer, pasad2023comparative, pasad2024self, choi2024self} and neuron-wise~\cite{lin2024property, wu2024and} analyses of acoustic, linguistic, and speaker properties. There are also studies analyzing supervised models like speech recognition~\cite{yang2024prompts, ngueajio2022hey} and emotion recognition~\cite{lin2024emo}. In contrast, existing work on LALMs focuses on high-level behaviors like bias~\cite{lin2024listen} and hallucination~\cite{leng2024curse}, lacking the internal analysis seen in SSL models. This motivates us to move beyond macroscopic observations and examine how auditory information is represented inside LALMs.
\begin{figure}[t]
    \centering

    \includegraphics[width=0.99\linewidth]{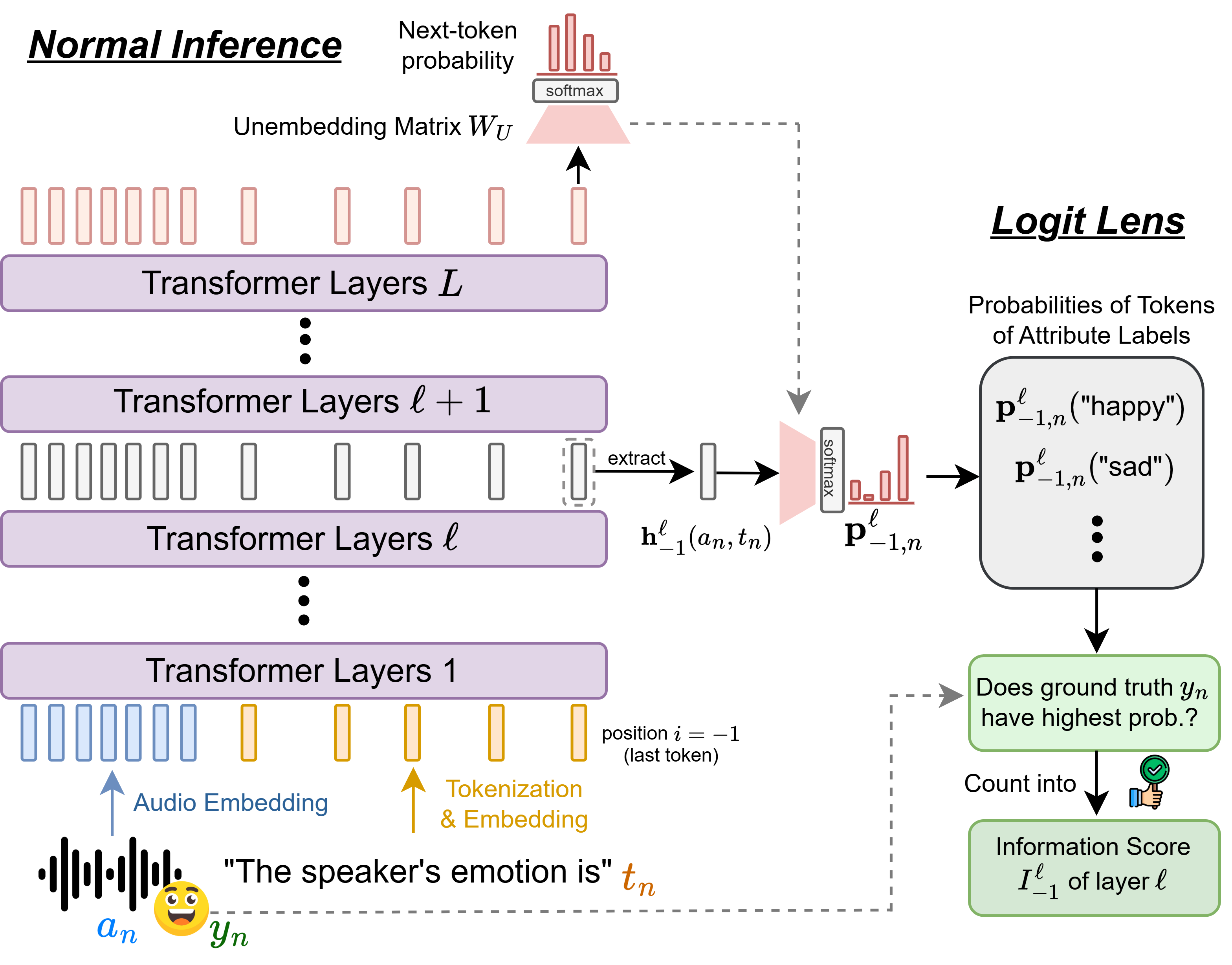}
    \vspace{-5pt}
    \caption{Illustration of Logit Lens and our method for investigating the internal evolution of attribute information in LALMs based on it.}
    \label{fig:illu}
\end{figure}
\subsection{Interpretability Methods for LLMs}

Understanding models' internal mechanisms is crucial for interpretation and improvement. As LALMs extend LLMs with auditory capabilities, we leverage interpretability techniques proven effective for LLMs and multimodal models. Specifically, common approaches analyze attention patterns~\cite{kobayashi2020attention}, neuron activations~\cite{tang2024language, rai2024investigation, zhao2025understanding}, or hidden representations~\cite{logit_lens, din2024jump, patchscopes}, and fall into training-based and training-free categories. Training-based methods use auxiliary modules like probing classifiers~\cite{belinkov2022probing}, while training-free methods analyze internal states during inference. Examples include identifying causal neurons via intervention~\cite{meng2022locating} and patching hidden representations to trace information flow~\cite{patchscopes}. We adopt Logit Lens~\cite{logit_lens}, a training-free vocabulary projection method~\cite{geva2022transformer, din2024jump, tunedlens}, for effective interpretation without extra training. We introduce this method in Sec.~\ref{sec:preliminary}.


\section{Problem Formulation}
\label{sec:formulation}
We investigate LALMs' internal behavior when perceiving and recognizing auditory attributes from sound inputs. Specifically, we address the following research questions (RQs):

\begin{enumerate}
\item \label{rq1} How does attribute information evolve across layers?
\item \label{rq2} Does this evolution differ between successful and unsuccessful attribute recognition? If so, how?
\item \label{rq3} At which layer do LALMs resolve attribute information, and does it correlate with recognition accuracy?
\item \label{rq4} How does auditory attribute information flow across token positions when recognizing attributes? 
\item \label{rq5} How can we improve LALMs with the above analyses?
\end{enumerate}

These questions explore how auditory attribute information is processed across layers and token positions in LALMs. By comparing its evolution in successful and failed recognition and identifying the typical resolution layer, we clarify the dynamics of attribute recognition. Understanding information flow across token positions elucidates how LALMs use internal information at different positions, including auditory inputs and preceding text tokens. These analyses lay the groundwork for interpreting model behavior and informing future improvements.

\section{Methods}

\subsection{Preliminaries: Logit Lens}
\label{sec:preliminary}
Logit Lens~\cite{logit_lens} is a simple yet powerful way to study what a language model ``knows" at each layer and token position. By projecting hidden representations back onto the vocabulary space, one can determine which tokens the model implicitly favors, revealing encoded facts, attributes, and relationships~\cite{yang2024large, geva2022transformer, merullo2024language}. It has proven to be an effective and valuable tool for interpreting text LLMs~\cite{geva2023dissecting, yang2024large, geva2022transformer, merullo2024language, biran2024hopping, logit_lens, wang2025logitlens4llms} and multimodal models~\cite{jiang2024interpretingeditingvisionlanguagerepresentations, neo2024towards, huo2024mmneuron}. We introduce this technique.

Consider an LLM with $L$ layers, hidden dimension $d$, and vocabulary $V$ of size $|V|$. To examine the information at token position $i$ in layer $\ell$, let $\mathbf{h}^\ell_i\in\mathbb{R}^d$ denote the hidden representation at position $i$ and layer $\ell$. Logit Lens projects $\mathbf{h}^\ell_i$ onto the vocabulary space via the model’s unembedding matrix\footnote{The unembedding matrix is the language model (LM) head that maps the final-layer hidden representations to logits over the vocabulary, which are then converted into a probability distribution for next-token prediction.} $W_U$ $\in$ $\mathbb{R}^{|V|\times d}$, producing a vector of logits. Applying softmax yields a probability distribution over the vocabulary:
\begin{equation}
  \label{eq:vocabulary_projection}
  \mathbf{p}^\ell_i
  \;=\;
  \mathrm{softmax}\bigl(W_U\,\mathbf{h}^\ell_i\bigr)
  \;\in\;
  \mathbb{R}^{|V|}
\end{equation}
The resulting distribution \(\mathbf{p}^\ell_i\) reflects the model's implicit preference for tokens at the given layer and position, thus serving as a basis for analyzing encoded information. An overview of this process is illustrated in Fig.~\ref{fig:illu}.

The effectiveness of the Logit Lens technique stems from the residual stream in transformer models, where each layer adds information into the stream and promotes the probability of concepts it encodes~\cite{geva2022transformer}. Prior studies have shown that these intermediate probability distributions encode rich and interpretable internal information, including factual knowledge, attributes, and relational cues about entities~\cite{yang2024large, geva2022transformer, merullo2024language}. For example, if position $i$ corresponds to the last token in a description of an entity $e$\footnote{For entities that span multiple tokens, a common practice is to use the first token as a representative~\cite{yang2024large}.}, then the probability $p^\ell_i(e)$ assigned to $e$ can serve as a proxy for how much information about \(e\) is recoverable at layer \(\ell\) when processing the description~\cite{yang2024large}. Note that the distribution $\mathbf{p}^L_i$ at the final layer $L$ matches the next-token probability distribution at position $i$.

Building upon this framework, we leverage intermediate layer distributions to quantify each layer’s contribution to encoding auditory attribute information. Specifically, we define a layer-wise information score to measure this encoding and identify critical layers where attribute resolution occurs. Based on these, we conduct analyses addressing the RQs in Sec.~\ref{sec:formulation}.


\subsection{Layer-wise Information Score}
\label{sec:information_score}

We first introduce the \textit{layer-wise information score} $I^\ell_i$, which measures how well the hidden representation at layer $\ell$ and token position $i$ of an LALM encodes auditory attribute information and resolves the attributes. An illustration of the layer-wise information score is included in Fig.~\ref{fig:illu}.

Given a dataset $\mathcal{D} = \{(a_n, t_n, y_n)\}_{n=1}^{|\mathcal{D}|}$, where $a_n$ is the audio input, $t_n$ the textual input, and $y_n$ the corresponding attribute label of $a_n$, and let $Y$ be the set of all attribute labels. For each $(a_n, t_n)$, the model produces a hidden representation $\mathbf{h}^\ell_i(a_n, t_n)$ at layer $\ell$ and token position $i$. We then define the layer-wise information score as:
\begin{equation}
\label{eq:layer_information}
I^\ell_i = \mathbb{E}_{(a_n, t_n, y_n)\in\mathcal{D}}\Bigl[\mathbb{I}\bigl(y_n = \mathop{\mathrm{argmax}}\limits_{y \in Y}\ \mathbf{p}^\ell_{i,n}(y)\bigr)\Bigr]
\end{equation}
Here, $\mathbb{I}(\cdot)$ is the indicator function (1 if the condition is true, 0 otherwise), and $\mathbf{p}^\ell_{i,n}$ is the probability distribution obtained by applying Eq. (\ref{eq:vocabulary_projection}) to $\mathbf{h}^\ell_i(a_n, t_n)$, with $\mathbf{p}^\ell_{i,n}(y)$ being the probability of the token of the attribute $y$ from this distribution.

Intuitively, $I^\ell_i$ can be viewed as the accuracy of predicting the attribute label from $\mathbf{h}^\ell_i$. A higher value of $I^\ell_i$ indicates that this layer’s representation not only captures the correct attribute but also boosts its probability above all other labels, thereby encoding more salient attribute information.

\subsection{Critical Layer Computation}

To capture where the model primarily resolves an auditory attribute at token position $i$, we compute a weighted average of layer indices, using each layer’s contribution as the weight. This weighted average layer is defined as the \textit{critical layer}, which naturally summarizes how attribute information is distributed across layers and provides an estimate of where LALMs resolve these attributes.

Formally, we build on the layer-wise information scores $I_i^\ell$ introduced earlier. Since $I_i^\ell$ behaves like an accuracy with a chance-level baseline of $1/|Y|$, we consider a layer $\ell$ at position $i$ to contribute meaningful attribute information only if its information score exceeds a threshold $(1+\alpha)/|Y|$, where $\alpha > 0$. We define the contribution of layer $\ell$ as:
\begin{equation}
s^\ell_i = \max\left(0, I_i^\ell - \frac{1+\alpha}{|Y|}\right)
\end{equation}
with $\alpha=0.2$ in our experiments. This thresholding filters out layers whose information scores barely surpass chance level, thereby reducing noises in the layer-wise information scores and enhancing the robustness of our analysis.

The critical layer $\ell^*_i$ is computed as the weighted average of layer indices, weighted by their contributions:
\begin{equation}
\label{eq:critical_layer}
\ell^*_i = \frac{\sum_{\ell=1}^L s^\ell_i \cdot \ell}{\sum_{\ell=1}^L s^\ell_i}
\end{equation}

A larger $\ell^*_i$ indicates that attribute information is concentrated in deeper layers, implying later resolution.

\section{Experimental Setup}
\subsection{Dataset}

We focus on four auditory attributes: speaker gender, spoken language, speaker emotion, and the animal producing the sound. The dataset contains triplets comprising an audio input, a textual prompt, and the corresponding attribute label.  The audio samples and attribute labels are sourced from the SAKURA\footnote{\url{https://github.com/ckyang1124/SAKURA}} benchmark~\cite{sakura}, which provides 500 samples per attribute. There are 2, 8, 5, and 9 distinct labels for gender, language, emotion, and animal, respectively.

We use three distinct prompt formats for textual inputs to probe how attribute information emerges across layers. 
\begin{enumerate}
    \item \textbf{Direct Prompt (P1)}: Templates like ``The speaker’s gender is."
    \item \textbf{Question-answer (QA) prompt (P2)}: We prepend a user-style question before the direct prompt to simulate a conversational QA scenario.
    \item \textbf{Multiple-choice (MC) prompt (P3)}: We extend P2 by including a list of possible attribute labels after the question to simulate MCQA scenarios.
\end{enumerate}

The formats are summarized in Table~\ref{tab:prompts}. Specifically, we focus on hidden representations at the final token (``is"). We choose this position because the model’s next token is highly likely to be the attribute itself, making it necessary to resolve the attribute by then. By measuring the layer-wise information scores, we identify layers reliably encoding the attribute.

\begin{table*}[t]\small 
\setlength\tabcolsep{2pt} 
\renewcommand{\arraystretch}{0.5}
\caption{Textual prompts used for different attributes and different prompt formats. P1, P2, and P3 denote the direct, QA, and MC prompt formats, respectively. \usermark and \asstmark represent tokens for headers that separate the turns in the models' chat templates.}

\centering

\resizebox{0.99\linewidth}{!}{
\begin{tabular}{c|c|c|c|c} 

\toprule
 & Gender & Language & Emotion & Animal \\

\midrule
P1 (Direct) & \makecell[c]{\asstmark The speaker's gender is} & \makecell[c]{\asstmark The speech's spoken language is} & \makecell[c]{\asstmark The speaker's emotion is} & \makecell[c]{\asstmark The sound file's animal is} \\

\midrule

P2 (QA) & \makecell[c]{\usermark What is the gender of the \\speaker in the speech?\asstmark \\The speaker's gender is} & \makecell[c]{\usermark What is the language spoken in \\the speech? \asstmark The speech's spoken \\language is} & \makecell[c]{\usermark What is the emotion of the \\speaker in the speech? \asstmark \\The speaker's emotion is} & \makecell[c]{\usermark What animal makes the \\sound? \asstmark The sound file's \\animal is} \\

\midrule

P3 (MC) & \makecell[c]{\usermark What is the gender of the \\speaker in the speech? Possible \\options: male, female. \asstmark \\The speaker's gender is} & \makecell[c]{\usermark What is the language spoken in the\\ speech? Possible options: English, German, Spanish,\\ French, Italian, Chinese, Japanese, Korean.\\ \asstmark The speech's spoken language is} & \makecell[c]{\usermark What is the emotion of the \\speaker in the speech? Possible options: \\angry, disgust, fear, happy, sad.\\ \asstmark The speaker's emotion is} & \makecell[c]{\usermark What animal makes the \\sound? Possible options: dog, cat, pig, \\cow, frog, hen, rooster, sheep, crow.\\ \asstmark The sound file's animal is}\\

\bottomrule

\end{tabular}
}
\label{tab:prompts}
\end{table*}

\subsection{Investigated Models}

We investigate three open-source LALMs: DeSTA2\footnote{\url{https://github.com/kehanlu/DeSTA2}}~\cite{desta2}, Qwen-Audio-Chat\footnote{\url{https://github.com/QwenLM/Qwen-Audio}} (Qwen)~\cite{qwen}, and Qwen2-Audio-Instruct\footnote{\url{https://github.com/QwenLM/Qwen2-Audio}} (Qwen2)~\cite{qwen2}. These models are selected for their strong performance on the attribute recognition tracks of SAKURA~\cite{sakura}, from which we source our dataset. Additionally, they perform competitively on other speech and audio benchmarks~\cite{airbench, mmau}, making them well-suited for our analyses. We implement Logit Lens on these models based on the Patchscopes toolkit\footnote{\url{https://pair-code.github.io/interpretability/patchscopes/}}~\cite{patchscopes}.



\section{Results}

\subsection{RQ1: Attribute Information Evolution Across Layers}
\label{sec:rq1}
\captionsetup[subfigure]{skip=0pt}
\begin{figure*}[t]
    \centering

    \subfloat[DeSTA2 on Gender\label{fig:rq1_desta2_gender}]{
        \includegraphics[width=0.235\textwidth, height=1.75cm]{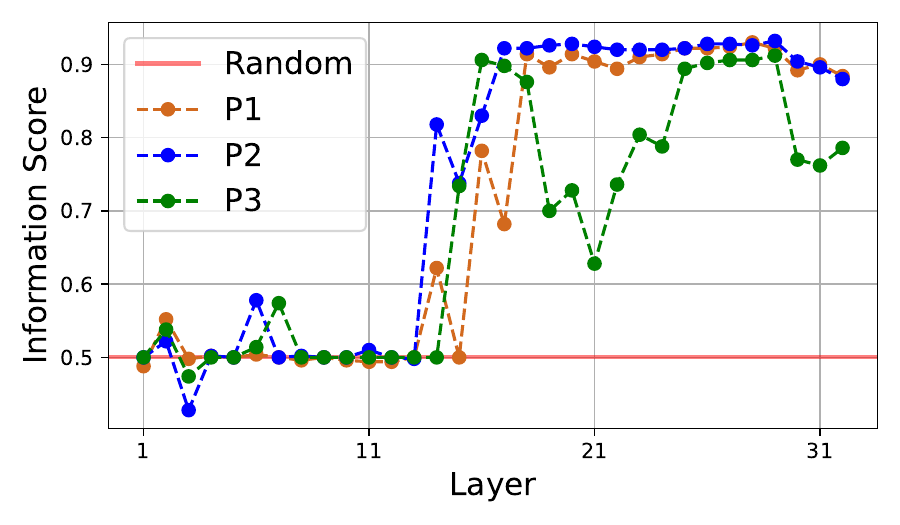}
    }\hfil
    \subfloat[DeSTA2 on Language\label{fig:rq1_desta2_language}]{
        \includegraphics[width=0.235\textwidth, height=1.75cm]{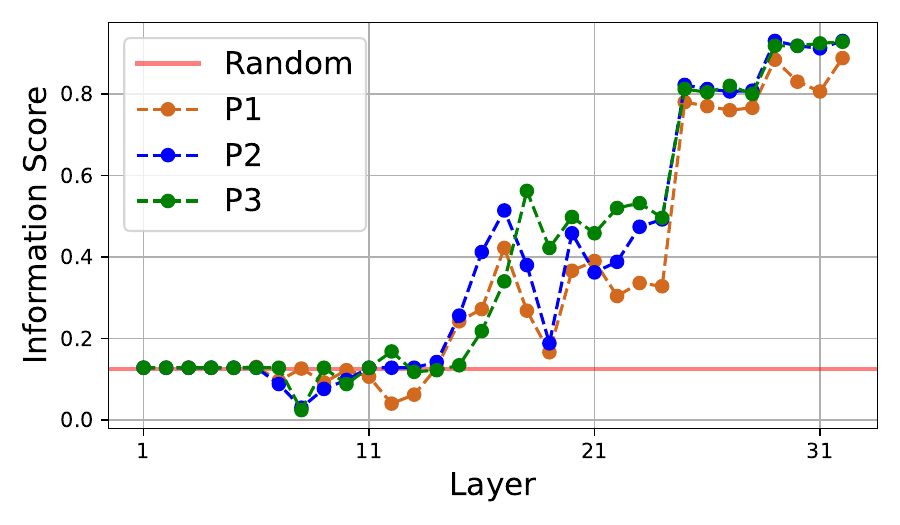}
    }\hfil
    \subfloat[DeSTA2 on Emotion\label{fig:rq1_desta2_emotion}]{
        \includegraphics[width=0.235\textwidth, height=1.75cm]{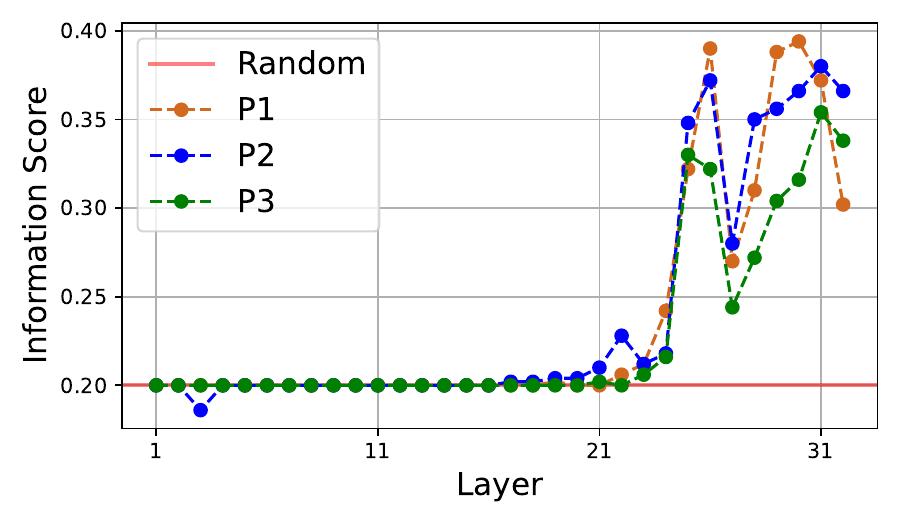}
    }\hfil
    \subfloat[DeSTA2 on Animal\label{fig:rq1_desta2_animal}]{
        \includegraphics[width=0.235\textwidth, height=1.75cm]{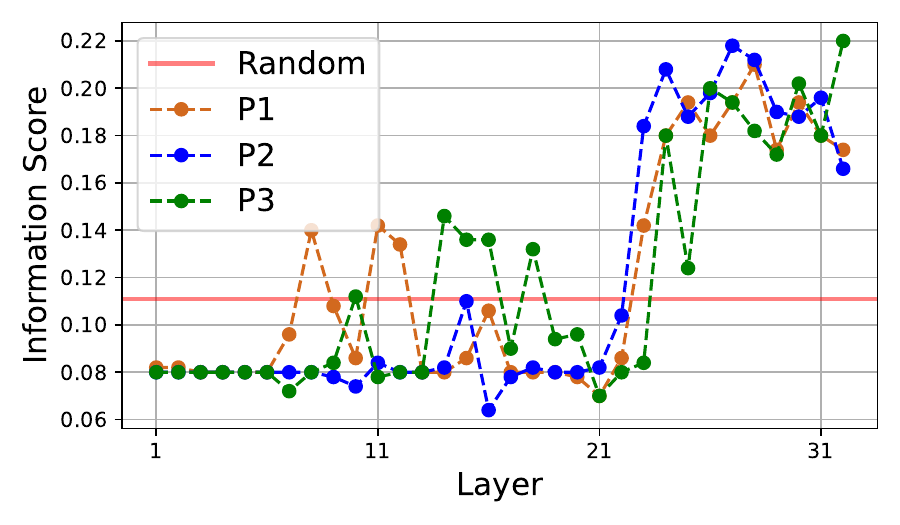}
    }\\[-8pt]

    \subfloat[Qwen on Gender\label{fig:rq1_qwen_gender}]{
        \includegraphics[width=0.235\textwidth, height=1.75cm]{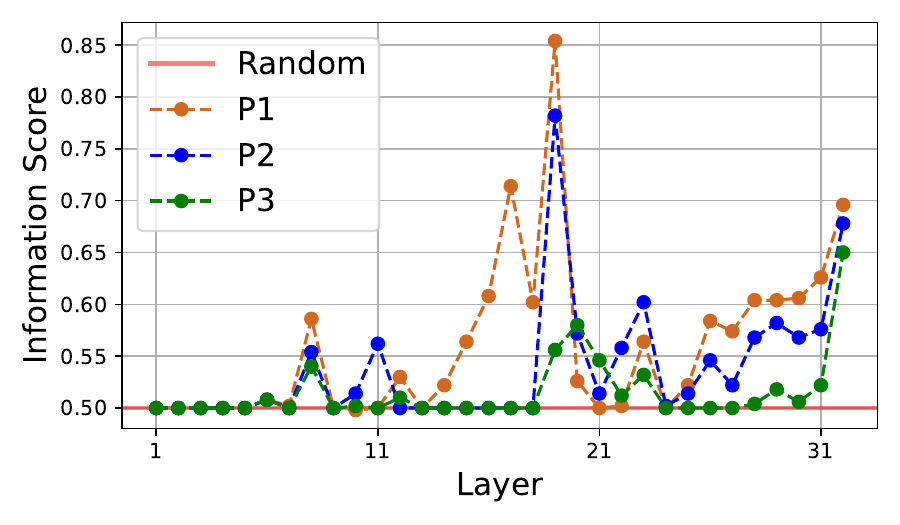}
    }\hfil
    \subfloat[Qwen on Language\label{fig:rq1_qwen_language}]{
        \includegraphics[width=0.235\textwidth, height=1.75cm]{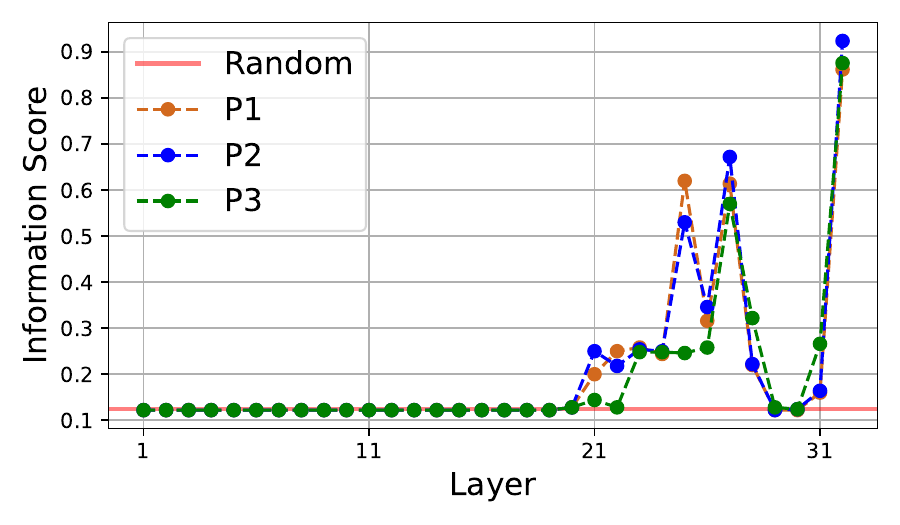}
    }\hfil
    \subfloat[Qwen on Emotion\label{fig:rq1_qwen_emotion}]{
        \includegraphics[width=0.235\textwidth, height=1.75cm]{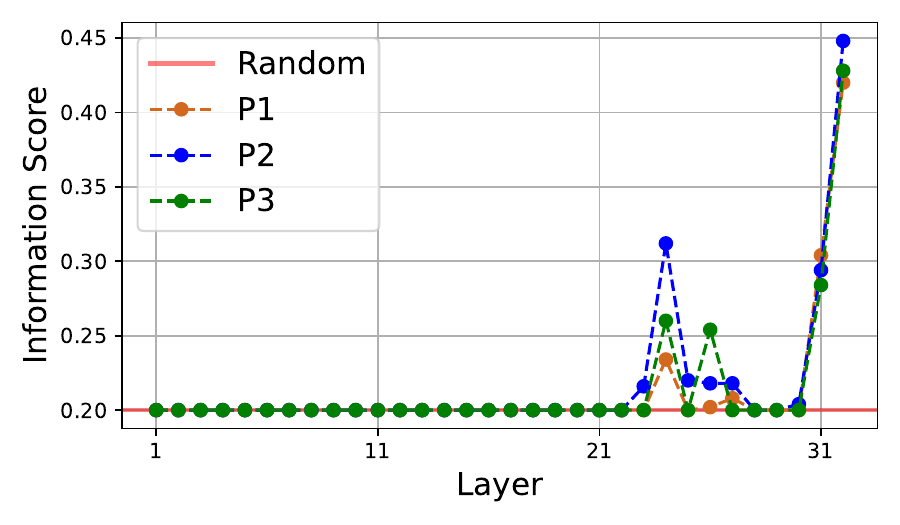}
    }\hfil
    \subfloat[Qwen on Animal\label{fig:rq1_qwen_animal}]{
        \includegraphics[width=0.235\textwidth, height=1.75cm]{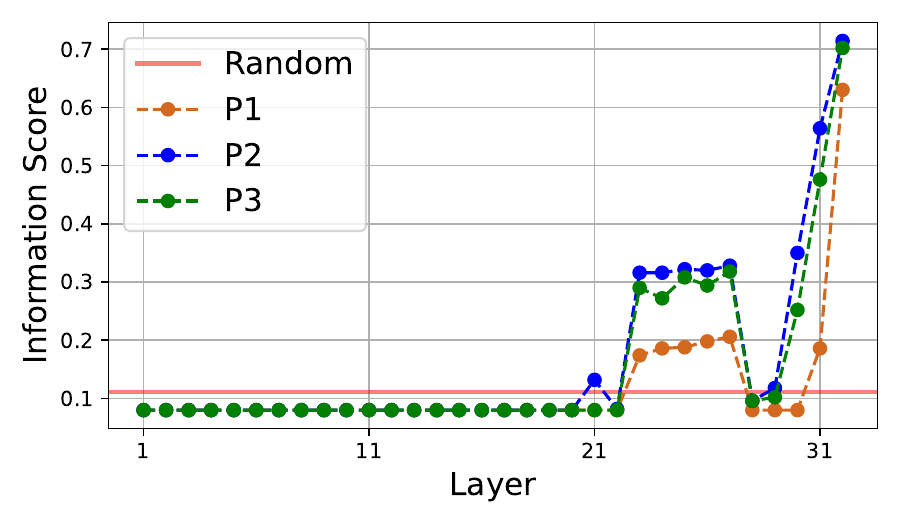}
    }\\[-8pt]

    \subfloat[Qwen2 on Gender\label{fig:rq1_qwen2_gender}]{
        \includegraphics[width=0.235\textwidth, height=1.75cm]{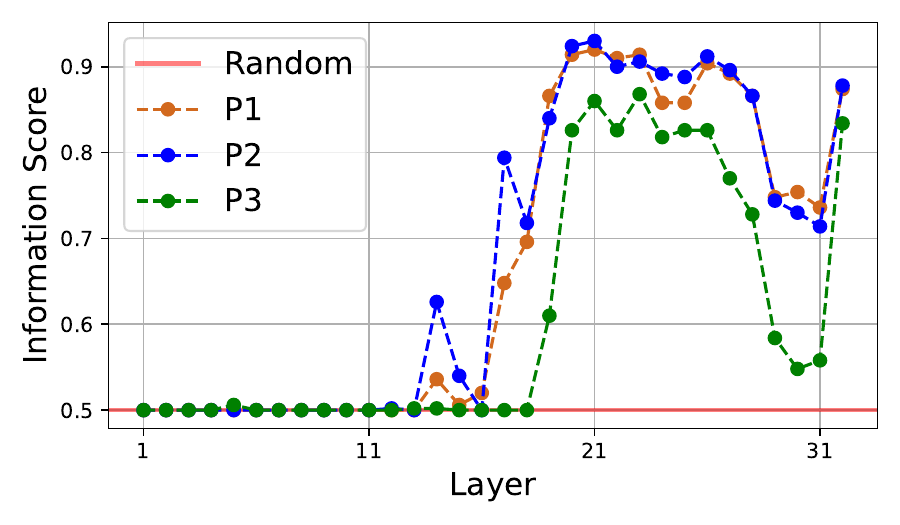}
    }\hfil
    \subfloat[Qwen2 on Language\label{fig:rq1_qwen2_language}]{
        \includegraphics[width=0.235\textwidth, height=1.75cm]{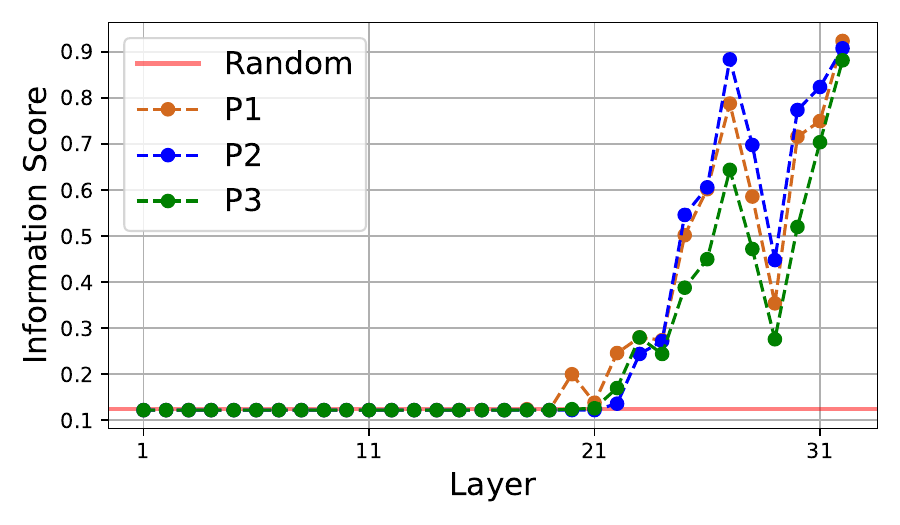}
    }\hfil
    \subfloat[Qwen2 on Emotion\label{fig:rq1_qwen2_emotion}]{
        \includegraphics[width=0.235\textwidth, height=1.75cm]{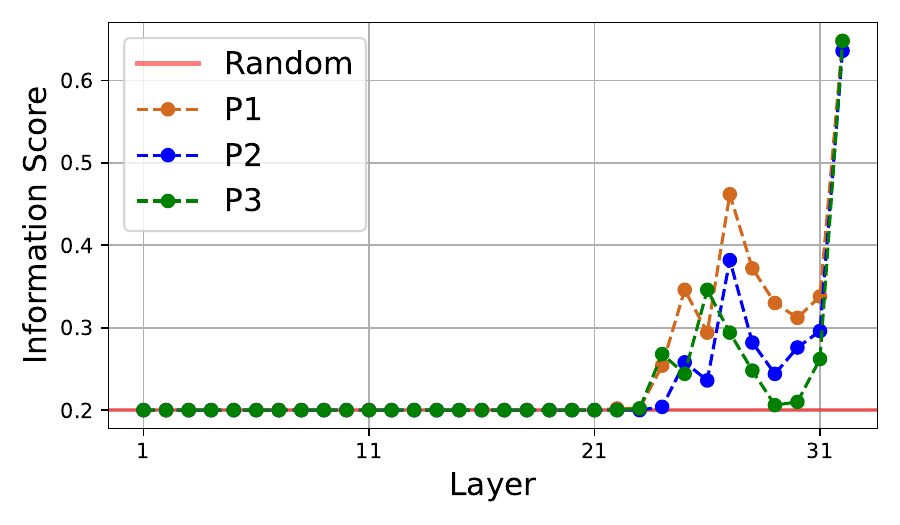}
    }\hfil
    \subfloat[Qwen2 on Animal\label{fig:rq1_qwen2_animal}]{
        \includegraphics[width=0.235\textwidth, height=1.75cm]{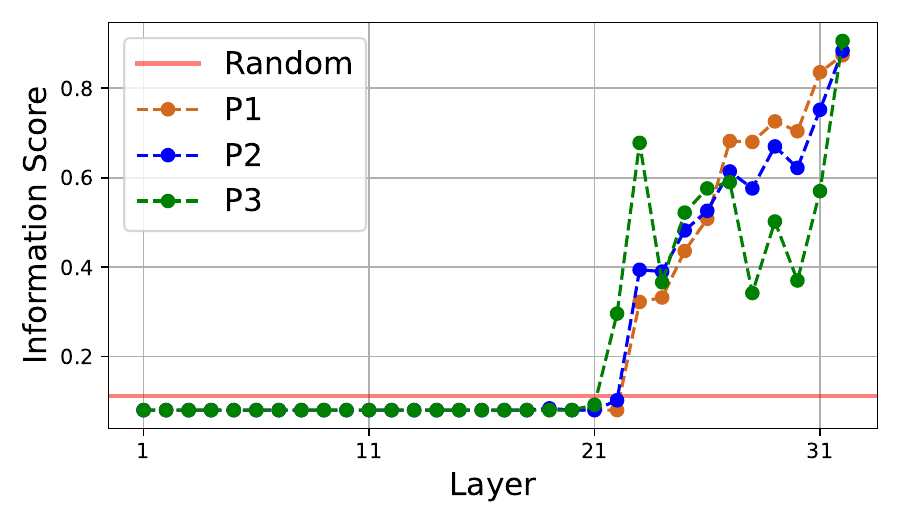}
    }\hfil

    \caption{Layer-wise information scores at the last token position of P1, P2, and P3, computed across all layers for three LALMs and four auditory attributes.}
    \label{fig:rq1}
\end{figure*}
We begin by addressing RQ1, investigating how auditory attribute information is represented across LALM layers. We compute the layer-wise information score at the last token (the token for ``is") under three prompt formats, as defined in Sec.~\ref{sec:information_score} and denoted as $I^\ell_{-1}$. The results are in Fig.~\ref{fig:rq1}.

Our first observation is that layers with low scores are close to the random baseline, defined as the reciprocal of the number of attribute labels. This confirms that layers without meaningful representations produce near-random predictions. An exception is DeSTA2 on the animal track, where some layers fall well below this baseline, likely due to limited training on animal sounds, causing unreliable predictions.

Generally, information scores increase with depth but not monotonically, with fluctuations and sharp drops followed by recoveries at deeper layers. Some recoveries fail, such as those for Qwen on the gender track (Fig.~\ref{fig:rq1_qwen_gender}).

Fig.~\ref{fig:rq1} also shows which layers best encode specific attributes. For example, gender information exhibits a distinct pattern concentrated in the middle-to-late layers and declines outside this range in Qwen and Qwen2. This pattern is specific to gender and not observed for other attributes, highlighting a characteristic encoding of gender information in these models.

Finally, information patterns are generally consistent across prompt formats, demonstrating stability against prompt variation. Therefore, we focus on results with P3 in the following sections, as it simulates typical multiple-choice QA settings.

\subsection{RQ2: Information Evolution in Correct/Wrong Predictions}
\label{sec:rq2}
\captionsetup[subfigure]{skip=0pt}
\begin{figure*}[t]
    \centering

    \subfloat[DeSTA2 on Gender\label{fig:rq2_desta2_gender}]{
        \includegraphics[width=0.235\textwidth, height=1.75cm]{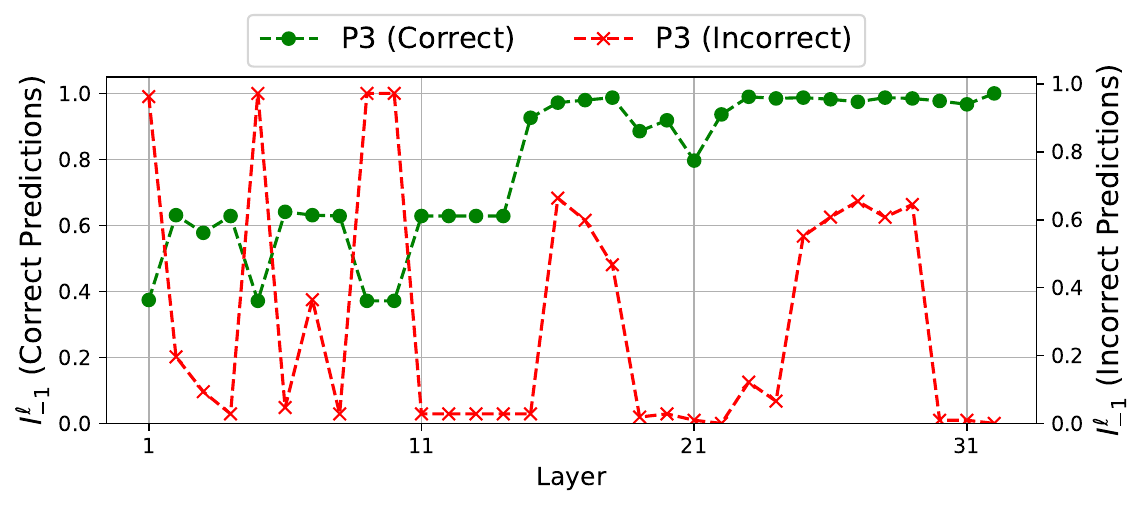}
    }\hfil
    \subfloat[DeSTA2 on Language\label{fig:rq2_desta2_language}]{
        \includegraphics[width=0.235\textwidth, height=1.75cm]{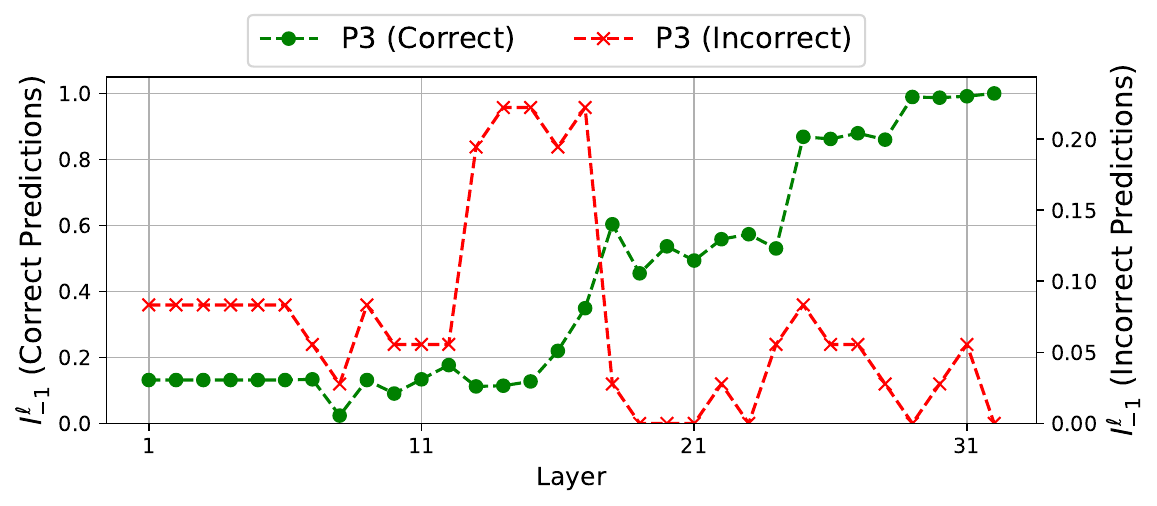}
    }\hfil
    \subfloat[DeSTA2 on Emotion\label{fig:rq2_desta2_emotion}]{
        \includegraphics[width=0.235\textwidth, height=1.75cm]{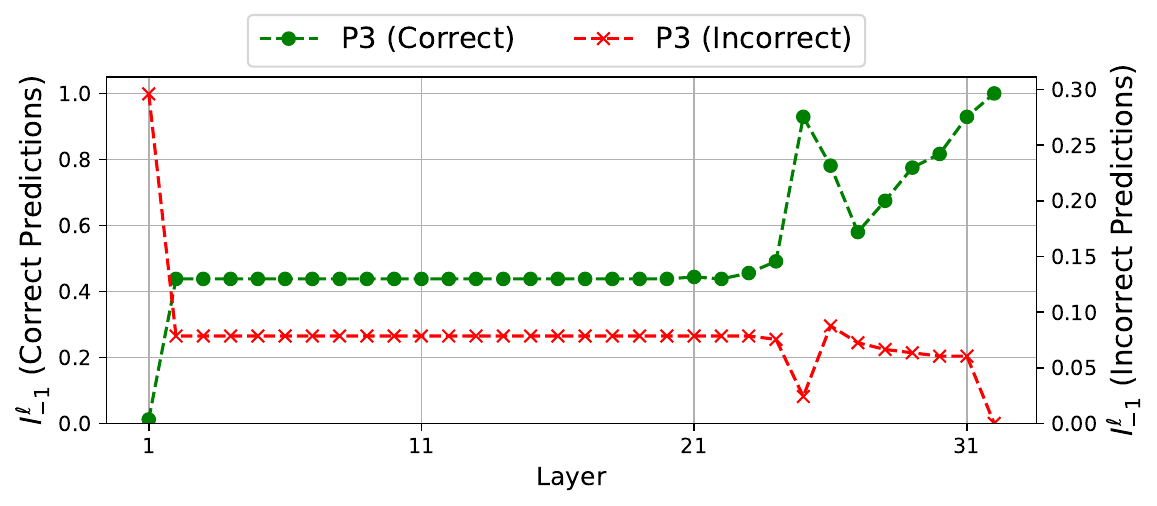}
    }\hfil
    \subfloat[DeSTA2 on Animal\label{fig:rq2_desta2_animal}]{
        \includegraphics[width=0.235\textwidth, height=1.75cm]{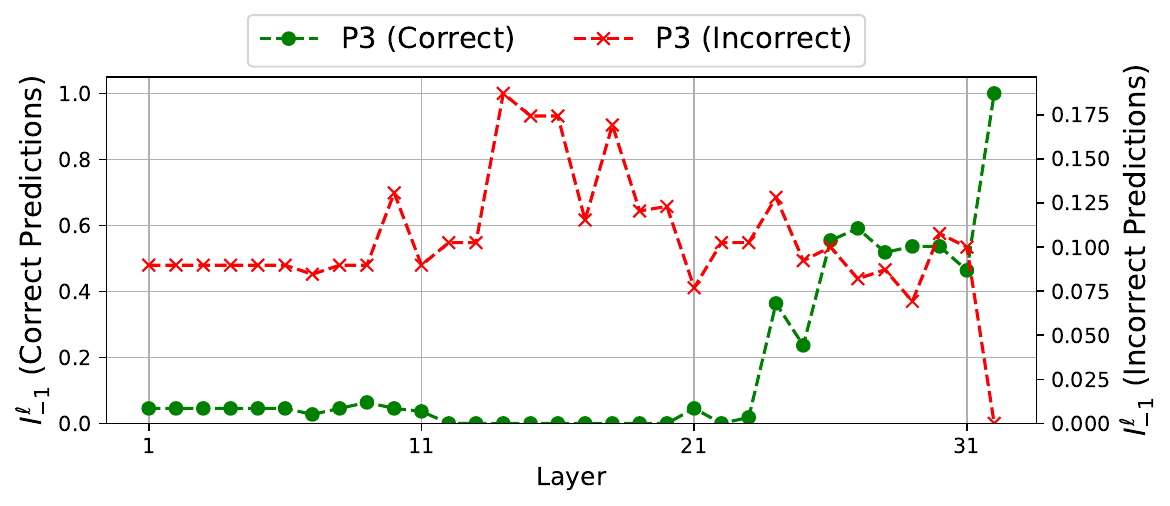}
    }\\[-8pt]

    \subfloat[Qwen on Gender\label{fig:rq2_qwen_gender}]{
        \includegraphics[width=0.235\textwidth, height=1.75cm]{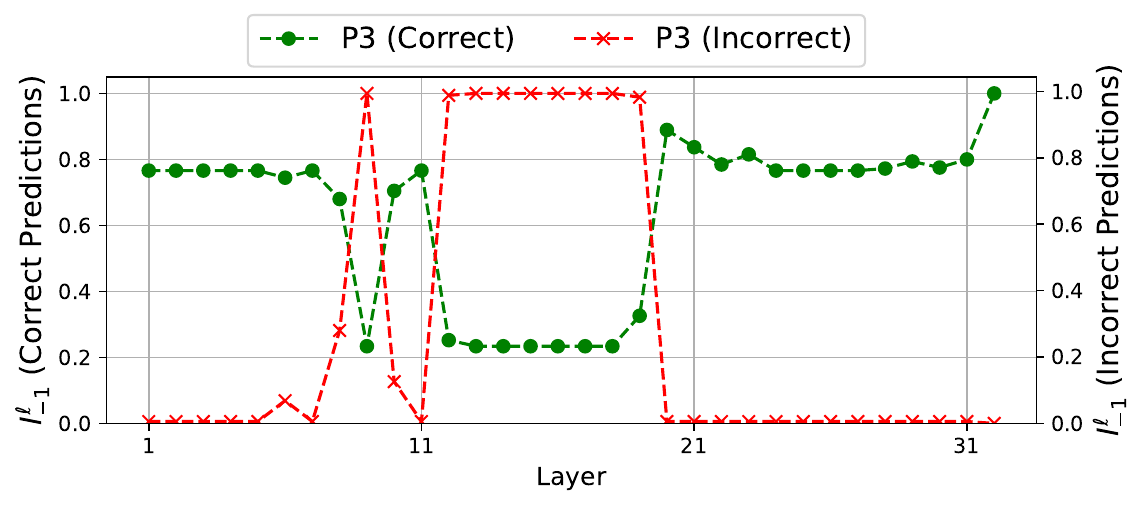}
    }\hfil
    \subfloat[Qwen on Language\label{fig:rq2_qwen_language}]{
        \includegraphics[width=0.235\textwidth, height=1.75cm]{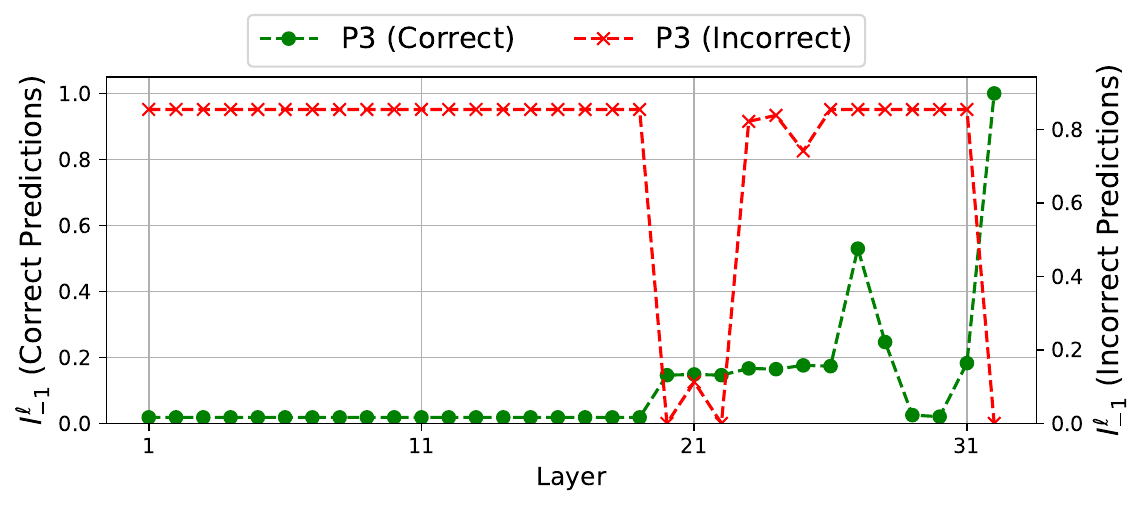}
    }\hfil
    \subfloat[Qwen on Emotion\label{fig:rq2_qwen_emotion}]{
        \includegraphics[width=0.235\textwidth, height=1.75cm]{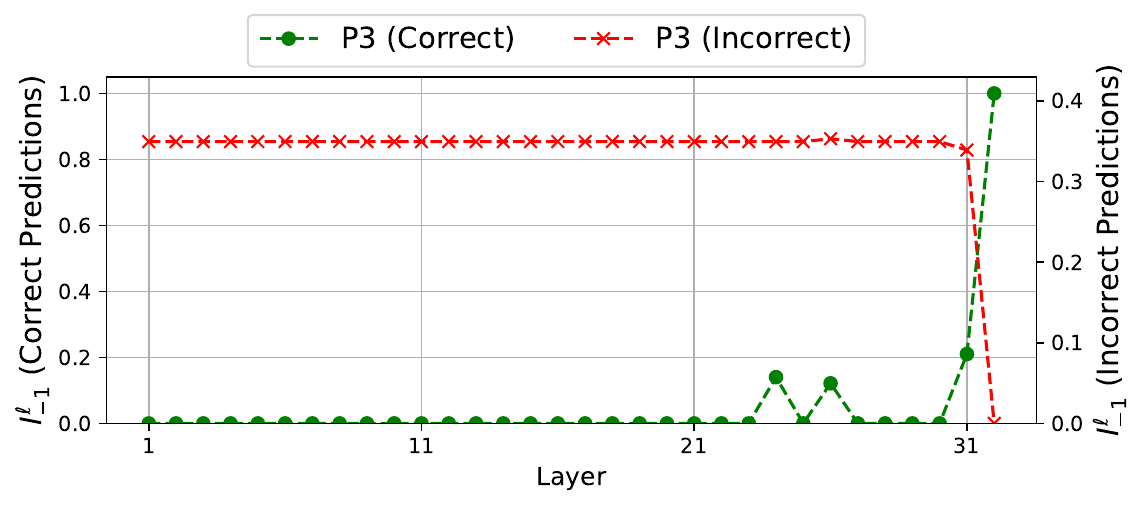}
    }\hfil
    \subfloat[Qwen on Animal\label{fig:rq2_qwen_animal}]{
        \includegraphics[width=0.235\textwidth, height=1.75cm]{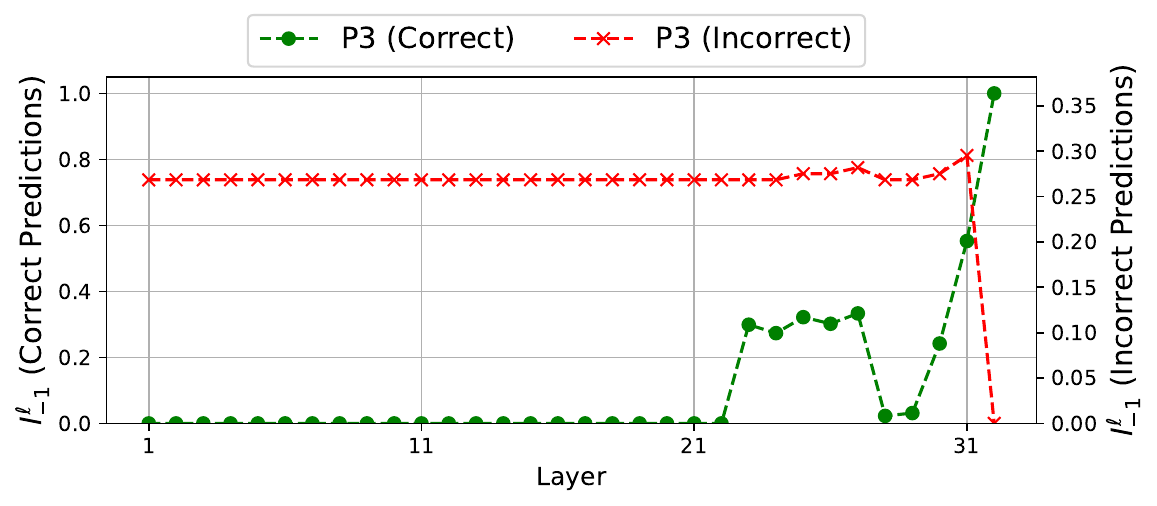}
    }\\[-8pt]

    \subfloat[Qwen2 on Gender\label{fig:rq2_qwen2_gender}]{
        \includegraphics[width=0.235\textwidth, height=1.75cm]{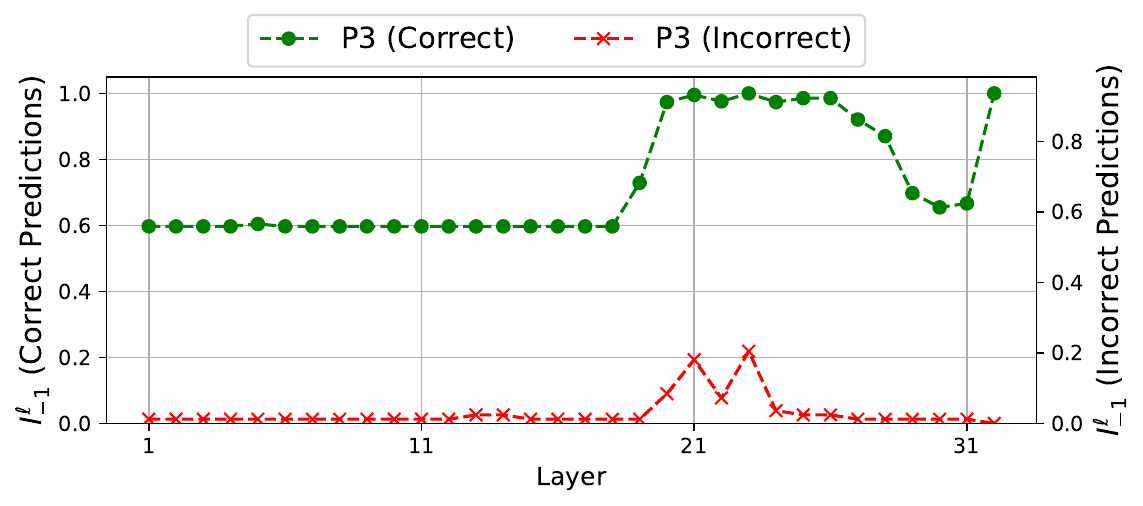}
    }\hfil
    \subfloat[Qwen2 on Language\label{fig:rq2_qwen2_language}]{
        \includegraphics[width=0.235\textwidth, height=1.75cm]{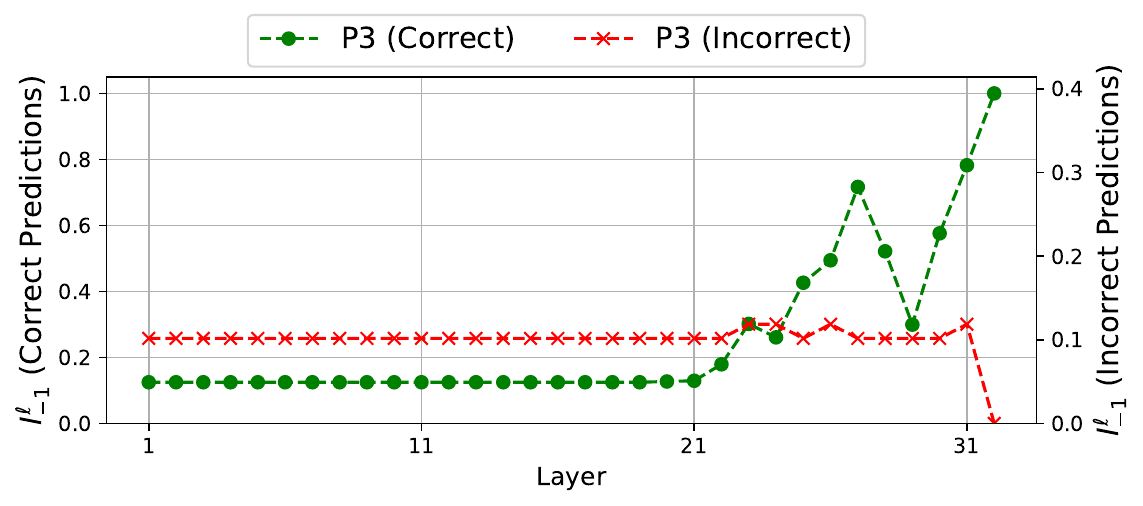}
    }\hfil
    \subfloat[Qwen2 on Emotion\label{fig:rq2_qwen2_emotion}]{
        \includegraphics[width=0.235\textwidth, height=1.75cm]{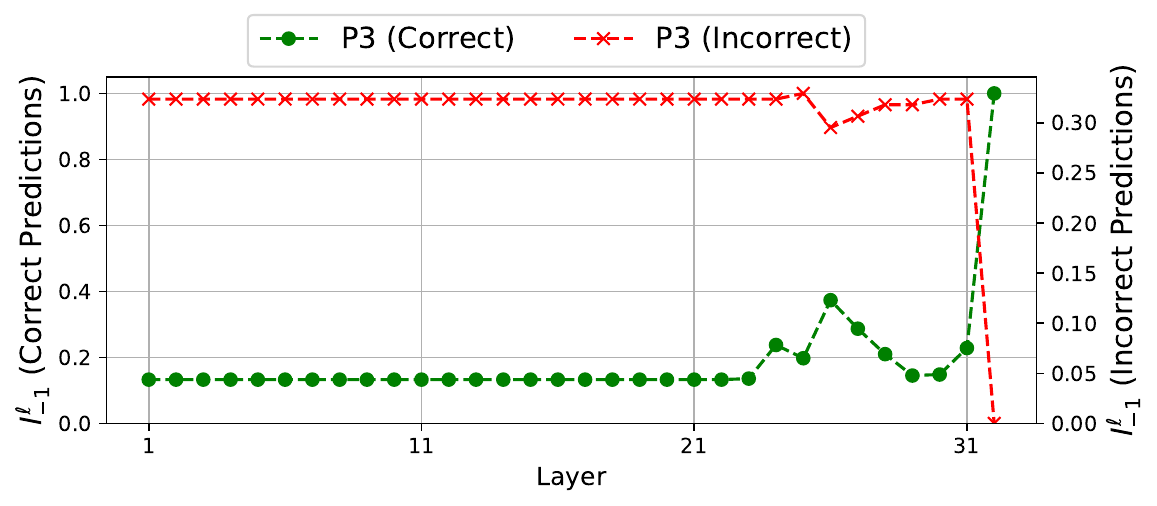}
    }\hfil
    \subfloat[Qwen2 on Animal\label{fig:rq2_qwen2_animal}]{
        \includegraphics[width=0.235\textwidth, height=1.75cm]{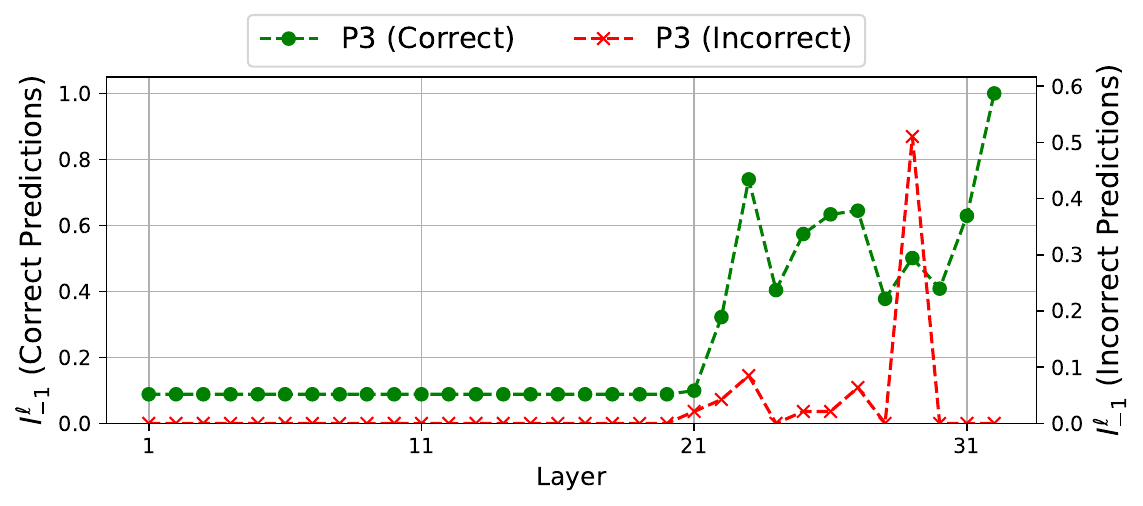}
    }\hfil

    \caption{Layer-wise information scores for three LALMs and four auditory attributes at the final token under P3 prompts. Green lines and left y-axis show scores for correctly predicted data; red lines and right y-axis show scores for incorrectly predicted data.}
    \label{fig:rq2}
\end{figure*}

We analyze attribute information evolution for samples with correct or incorrect predictions. A correct prediction means the ground-truth label has the highest next-token probability at the last token position (i.e., ``is") where LALMs are signaled to make predictions. Accordingly, the model’s prediction accuracy equals its $I^L_{-1}$\footnote{This aligns with the common likelihood-based accuracy metric, which checks if the ground truth holds the highest likelihood among options~\cite{mmlu}.}, where $L$ is the number of layers. Accuracy results are discussed in the next section.

For each model and attribute, we split the dataset into correctly and incorrectly predicted subsets and compute the layer-wise information score $I^\ell_{-1}$ separately for each subset. Results under the P3 prompt format are shown in Fig.~\ref{fig:rq2}.

We observe two contrasting trends: for correctly predicted samples (green lines), attribute information generally increases with depth; for incorrect predictions (red lines), information peaks at certain layers and then sharply declines, suggesting that some layers encode information well, but later ones degrade it, causing prediction errors. The superposition of these opposing dynamics explains the fluctuations in Sec.~\ref{sec:rq1}.

\subsection{RQ3: The Layer at Which LALMs Resolve Attribute Information and Its Correlation with Recognition Accuracy}
\label{sec:rq3}

In Sec.~\ref{sec:rq1} and~\ref{sec:rq2}, we examined how attribute information evolves across LALM layers. A natural question is whether this information evolution correlates with the models’ prediction accuracy. To investigate, we analyze the relationship between the attribute prediction accuracy, defined as $I^L_{-1}$ in Sec.~\ref{sec:rq2}, and the average layer where the attribute information is resolved, represented by the critical layers from Eq. (\ref{eq:critical_layer}). Table~\ref{tab:layers&accuracies} shows these values averaged over three prompt formats for the three LALMs.
\begin{table}[t]\small 
\setlength\tabcolsep{2pt} 
\renewcommand{\arraystretch}{0.2}
\footnotesize
\caption{Critical layers and accuracy (\%) of three LALMs on four attributes, averaged over three prompt formats. Values are shown as ``critical layer / accuracy".}
\centering

\resizebox{0.99\linewidth}{!}{
\begin{tabular}{c|c|c|c|c} 

\toprule
 & Gender & Language & Emotion & Animal \\
\midrule
DeSTA2 & 23.90 / 85.00 &	26.23 / 91.53 &	28.76 / 33.53 &	27.53 / 18.67 \\
\midrule

Qwen & 25.57 / 67.47 &	27.95 / 88.73 &	30.92 / 43.20 &	28.95 / 68.20 \\
\midrule

Qwen2 & 24.42 / 86.20 &	28.56 / 90.47 &	29.88 / 64.40 &	28.18 / 88.80 \\
\bottomrule

\end{tabular}
}
\label{tab:layers&accuracies}
\end{table}

We find that higher accuracy tends to align with shallower critical layers, with gender information resolved at the earliest layers, followed by language and animal, and emotion resolved at the deepest layers. To further quantify this, we calculate the Pearson correlation between critical layers and accuracies across models and prompts, as shown in Table~\ref{tab:correlation}.

For DeSTA2, this trend is clear with a significant negative correlation. For Qwen and Qwen2, the trend holds for attributes other than gender as well, echoing the unique encoding pattern for gender information described in Sec.~\ref{sec:rq1}. We conclude that, generally, resolving attribute information at earlier layers leads to a higher accuracy, possibly because more subsequent layers are available to refine and utilize the resolved information for correct prediction.
\begin{table}[t]\small 
\setlength\tabcolsep{2pt} 
\renewcommand{\arraystretch}{0.05}
\footnotesize
\caption{Pearson correlation and p-value between accuracies and critical layers for three LALMs. Significant p-values ($<$0.05) are bolded. ``Excluding Gender" indicates correlations computed without gender track data.}

\centering

\resizebox{0.99\linewidth}{!}{
\begin{tabular}{c|c|c} 
\toprule
 & Pearson Correlation & P-value \\
\midrule

DeSTA2 & -0.748 & $\mathbf{5.19\times10^{-3}}$ \\
\midrule

Qwen & -0.413 & $1.83\times10^{-1}$ \\
\midrule

Qwen (Excluding Gender) & -0.924 & $\mathbf{3.68\times10^{-4}}$ \\
\midrule

Qwen2 & -0.490 & $1.06\times10^{-1}$ \\
\midrule

Qwen2 (Excluding Gender) & -0.879 & $\mathbf{1.82\times10^{-3}}$ \\

\bottomrule

\end{tabular}
}
\label{tab:correlation}
\end{table}

\captionsetup[subfigure]{skip=0pt}
\begin{figure*}[t]
    \centering

    \subfloat[DeSTA2 on Gender\label{fig:rq4_desta2_gender}]{
        \includegraphics[width=0.235\textwidth, height=2cm]{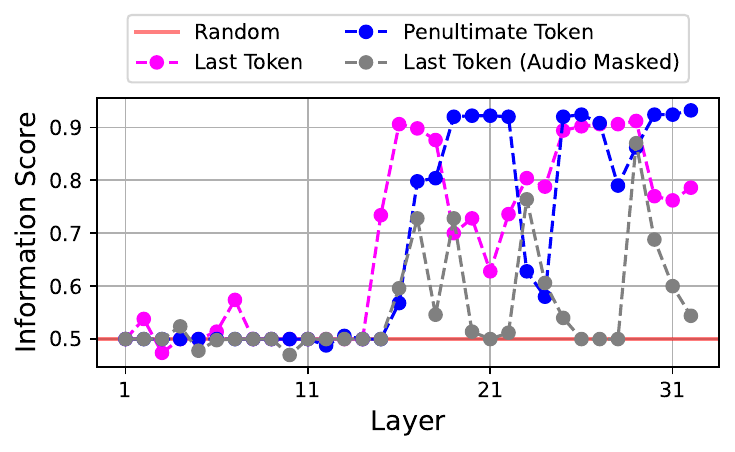}
    }\hfil
    \subfloat[DeSTA2 on Language\label{fig:rq4_desta2_language}]{
        \includegraphics[width=0.235\textwidth, height=2cm]{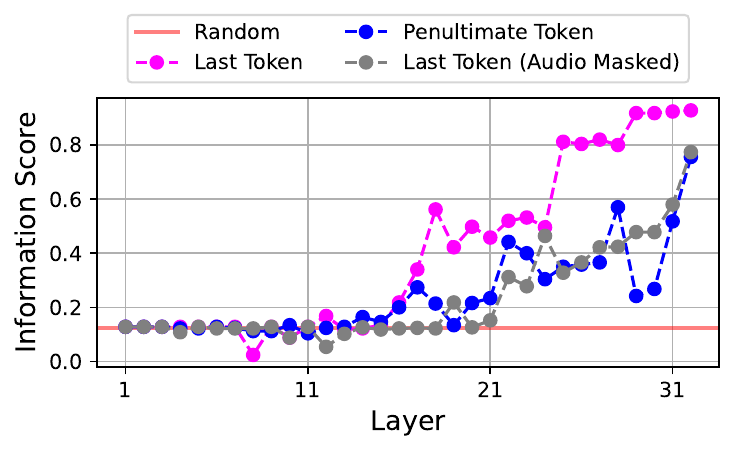}
    }\hfil
    \subfloat[DeSTA2 on Emotion\label{fig:rq4_desta2_emotion}]{
        \includegraphics[width=0.235\textwidth, height=2cm]{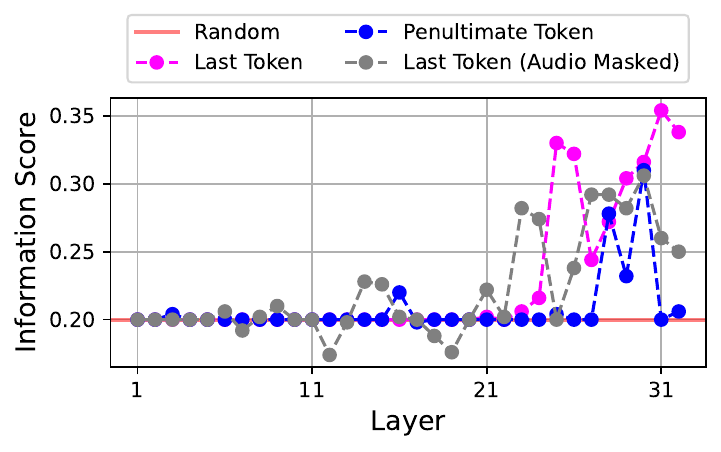}
    }\hfil
    \subfloat[DeSTA2 on Animal\label{fig:rq4_desta2_animal}]{
        \includegraphics[width=0.235\textwidth, height=2cm]{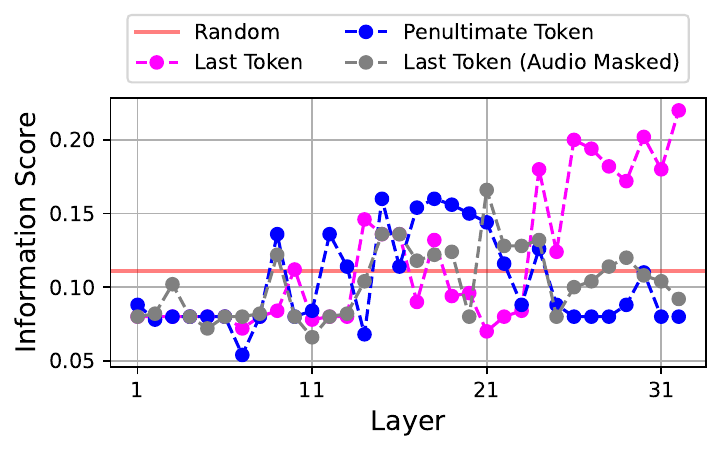}
    }\\[-8pt]

    \subfloat[Qwen on Gender\label{fig:rq4_qwen_gender}]{
        \includegraphics[width=0.235\textwidth, height=2cm]{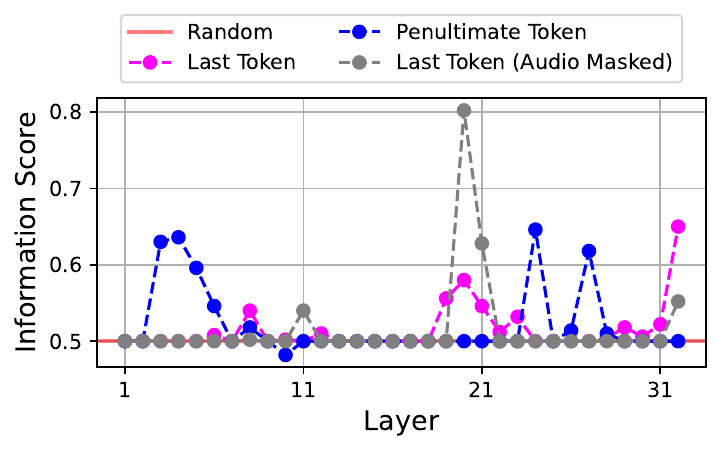}
    }\hfil
    \subfloat[Qwen on Language\label{fig:rq4_qwen_language}]{
        \includegraphics[width=0.235\textwidth, height=2cm]{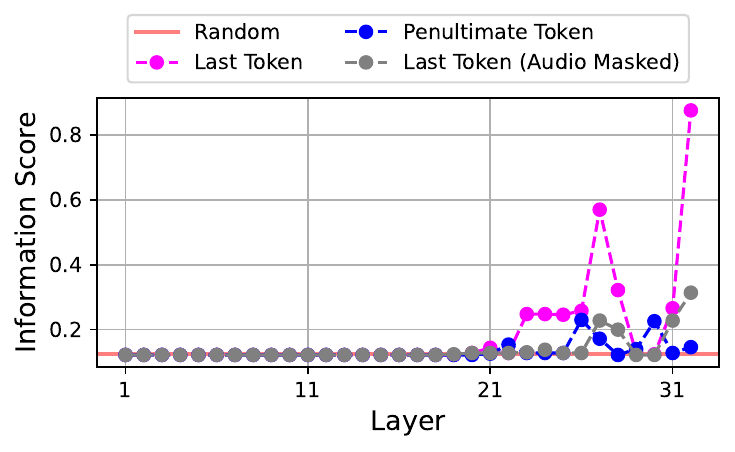}
    }\hfil
    \subfloat[Qwen on Emotion\label{fig:rq4_qwen_emotion}]{
        \includegraphics[width=0.235\textwidth, height=2cm]{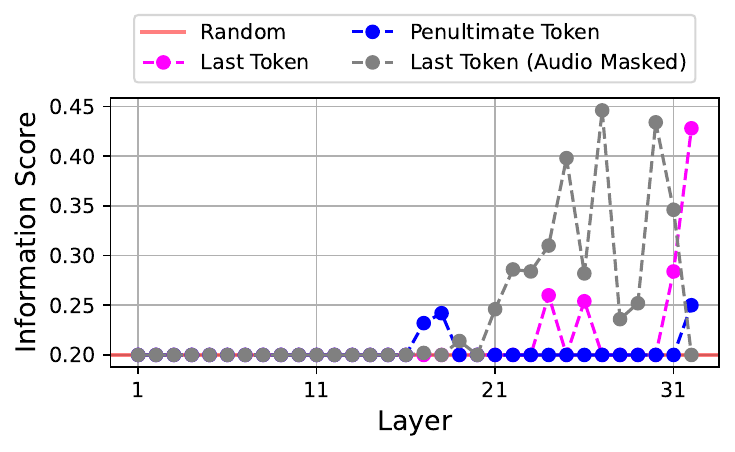}
    }\hfil
    \subfloat[Qwen on Animal\label{fig:rq4_qwen_animal}]{
        \includegraphics[width=0.235\textwidth, height=2cm]{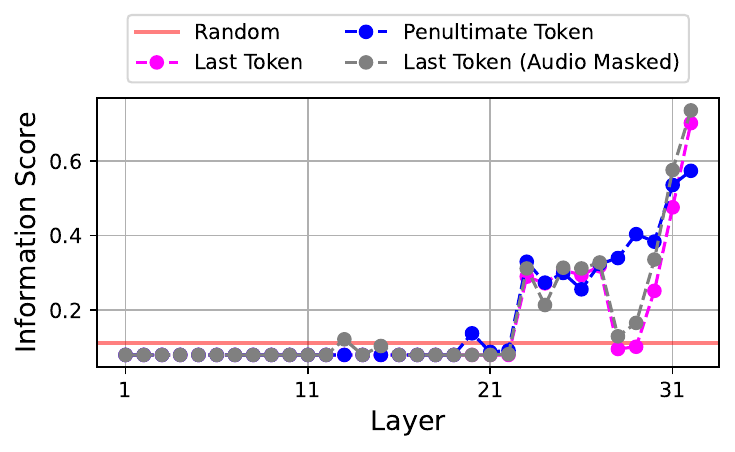}
    }\\[-8pt]

    \subfloat[Qwen2 on Gender\label{fig:rq4_qwen2_gender}]{
        \includegraphics[width=0.235\textwidth, height=2cm]{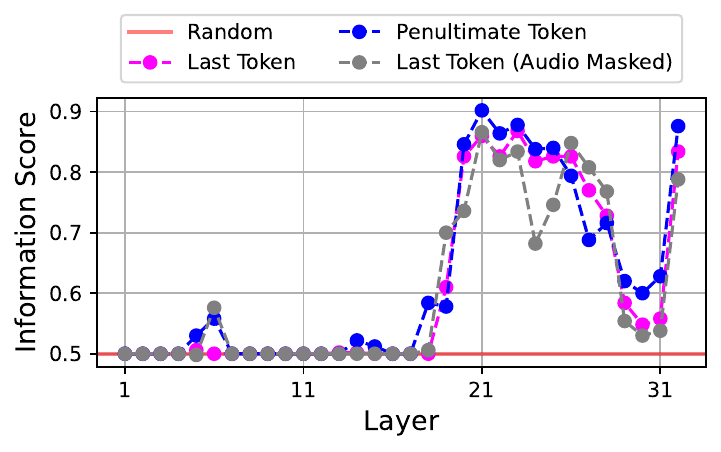}
    }\hfil
    \subfloat[Qwen2 on Language\label{fig:rq4_qwen2_language}]{
        \includegraphics[width=0.235\textwidth, height=2cm]{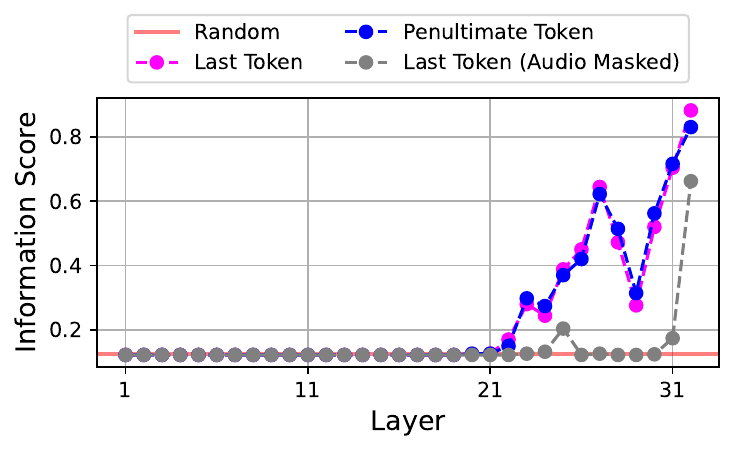}
    }\hfil
    \subfloat[Qwen2 on Emotion\label{fig:rq4_qwen2_emotion}]{
        \includegraphics[width=0.235\textwidth, height=2cm]{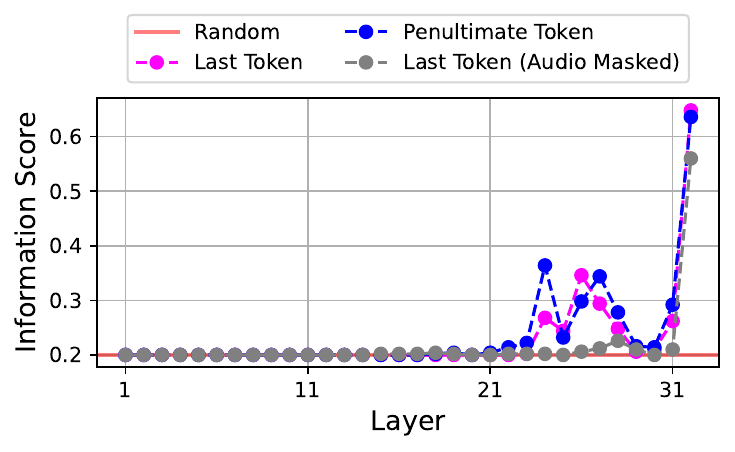}
    }\hfil
    \subfloat[Qwen2 on Animal\label{fig:rq4_qwen2_animal}]{
        \includegraphics[width=0.235\textwidth, height=2cm]{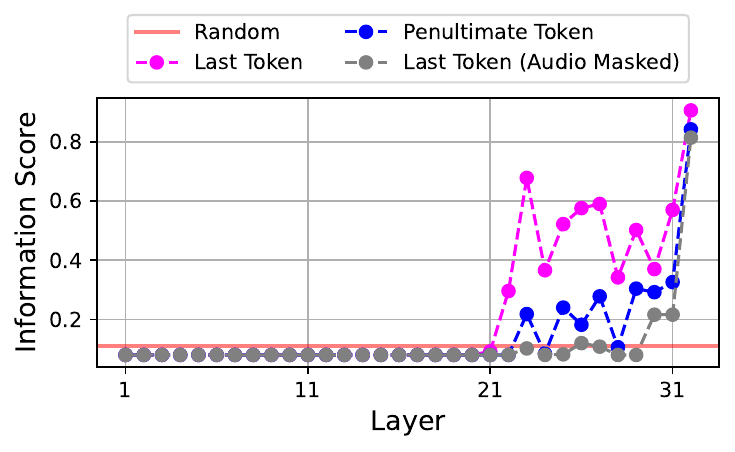}
    }\hfil

    \caption{Layer-wise information scores for three LALMs and four auditory attributes at the final token (i.e., the token “is”), the penultimate token (e.g., the token representing the attribute such as “gender”), and at the final token with auditory input positions masked during self-attention. Prompt format P3 is used.}
    \label{fig:rq4}

\end{figure*}
\subsection{RQ4: Information Flow Across Token Positions}
\label{sec:rq4}

In this section, we analyze how attribute information varies across token positions and identify the information sources LALMs rely on to predict attributes by comparing layer-wise information scores at two key token positions: the penultimate token, which explicitly mentions the attribute, and the last token, where LALMs make predictions. For example, in prompts like “The speaker’s gender is,” the penultimate token (“gender”) denotes the attribute, while the last token (“is”) signals the prediction. As the final token of the attribute mention, the hidden representation at the penultimate token is expected to contain essential attribute information~\cite{meng2022locating, geva2023dissecting, yang2024large}. Comparing these positions helps clarify how attribute information is encoded across token positions.

Fig.~\ref{fig:rq4} shows that, especially in the final few layers, information scores at the last token (pink lines) are typically higher than at the penultimate token (blue lines), with few exceptions, implying that LALMs are unlikely to rely solely on the hidden representations of preceding text tokens to make predictions.


To quantify this, we mask auditory inputs during self-attention\footnote{For DeSTA2, the inputs include speech transcriptions, which we treat as part of the auditory input. To ensure consistency with other models that do not use transcriptions, we mask them during processing.} at the last token\footnote{Masking applies only at the last token; other positions are unaffected.}, forcing the model to rely solely on hidden representations at preceding text token positions for attribute prediction (gray lines in Fig.~\ref{fig:rq4}). In most cases, we observe a notable drop in information scores and prediction accuracies, showing that information aggregated at the preceding text token positions alone is insufficient, and LALMs heavily rely on information directly obtained from auditory inputs when making predictions.

These findings have important implications for LALMs’ reasoning abilities. If the model fails to sufficiently consolidate relevant attribute information at attribute-mentioning positions and instead accumulates most of it when signaled to predict the attribute, it may struggle with reasoning requiring latent information integration. For example, multi-hop reasoning often lacks explicit cues that guide prediction (e.g., the last token ``is" in our prompts) at the attribute-mentioning positions, and insufficient early encoding can hinder subsequent reasoning. This aligns with prior work reporting limited multi-hop reasoning in LALMs~\cite{sakura}.

\subsection{RQ5: Demonstration of Applications}
\label{sec:rq5}

We present an example application demonstrating how our analyses can guide improvements in LALMs. As discussed in Sec.~\ref{sec:rq2}, attribute information across layers results from two opposing dynamics: increasing or decreasing with depth. Poor performance on recognizing certain attributes corresponds to the dominance of the decreasing dynamic.


Based on this, we hypothesize that enhancing deeper layer representations with information from earlier, richer layers could improve predictions. We conduct an experiment to verify the feasibility and effectiveness of this idea. Specifically, we split the dataset into two disjoint subsets: a probing set of 100 samples and a testing set of 400 samples. On the probing set, we compute layer-wise information scores at the last token to identify the layer $\bar{\ell}$ of highest attribute information among incorrectly predicted samples, serving as a proxy for where attribute information is most prominent in failure cases. Then, for each testing sample, we extract the hidden representation $\mathbf{h}^{\bar{\ell}}_{-1}$ at layer $\bar{\ell}$ and the last token and add it, scaled by a factor $\lambda$, to the representation five layers deeper:
\begin{equation}
    \mathbf{h}^{\bar{\ell}+5}_{-1} \leftarrow \mathbf{h}^{\bar{\ell}+5}_{-1} + \lambda \mathbf{h}^{\bar{\ell}}_{-1}
\end{equation}
The five-layer gap is chosen heuristically, as too small a gap may yield negligible enrichment, while too large a gap leaves insufficient subsequent layers to resolve the modification. We apply the same enrichment procedure to all testing samples.

\begin{figure}[t]
    \centering

    \includegraphics[width=0.99\linewidth]{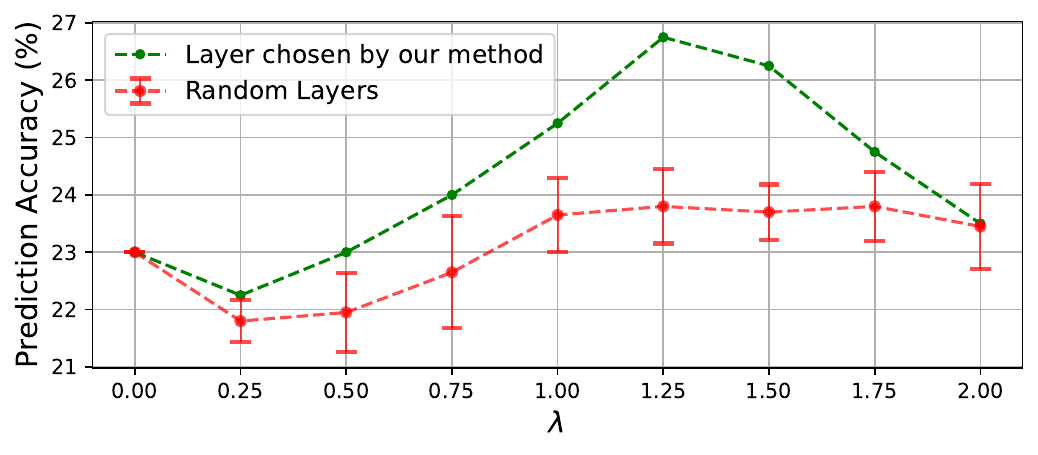}
    \vspace{-10pt}
    \caption{Accuracy (\%) of enriching the deeper layer using layers selected by our method versus random layers on a 400-sample test set. Random layer results are averaged over five seeds; error bars show standard deviation.}
    \label{fig:rq5}
\end{figure}
To demonstrate the effectiveness of this method in improving performance on challenging tasks, we present a representative case study on DeSTA2's animal recognition, an attribute that is especially challenging for DeSTA2, yielding the notably worst performance among all investigated models and attributes (see Table~\ref{tab:layers&accuracies}). The prompt format P3 is used.
We compare our method to a baseline where, for each sample, a random layer is selected as $\bar{\ell}$ for enrichment. This baseline is repeated five times with different random seeds.


Fig.~\ref{fig:rq5} shows accuracy on 400 testing samples across different $\lambda$ values. Our method, selecting $\bar{\ell}$ via layer-wise information scores, significantly outperforms the baseline over a wide range of $\lambda$, demonstrating its ability to identify layers containing meaningful information. We also observe that the choice of $\lambda$ is critical, as both excessively small and large values result in suboptimal performance. With a proper scaling factor $\lambda$ (i.e., when $\lambda=1.25$ in Fig.~\ref{fig:rq5}), our method achieves a relative accuracy improvement of 16.3\% over the original performance of DeSTA2 (i.e., when $\lambda=0$), without requiring any additional training.

This shows that selectively enriching deeper layers with information-rich earlier representations based on the layer-wise information scores improves performance. Our findings underscore the value of internal analysis for guiding model refinement and motivate future work on layer interaction and advanced enrichment methods to further enhance LALMs.

\section{Conclusion}

We present the first analysis of auditory attribute information evolution in LALMs across layers and token positions. We reveal two opposing dynamics: attribute information either increases or decreases with depth. Recognition failures occur when the latter dominates, where deeper layers degrade earlier encoded information. We find that resolving attribute information at earlier layers correlates with better accuracy. Token-wise analysis shows that information at attribute-mentioning positions alone is insufficient for attribute recognition; LALMs still rely heavily on directly querying auditory inputs. Finally, we demonstrate how these insights inform model improvement. Our work advances understanding of LALMs, laying a foundation for future research. Future work can explore strategies for improved information consolidation and layer interaction to further advance LALM capabilities.

\section{Limitations}

We acknowledge several limitations of this work. Our analysis of LALMs employs the Logit Lens technique, which is common in existing literature, though we note that alternative variants of this method exist. Furthermore, as the first study to examine the auditory information processing of LALMs, our investigation primarily focuses on their ability to perceive and recognize fundamental auditory attributes. While these capabilities form the foundation for more advanced reasoning over auditory modalities, the underlying reasoning processes of LALMs are likely to involve distinct, multi-layered, and potentially more intricate patterns than those observed in foundational attribute perception. A comprehensive examination of such processes, particularly how LALMs integrate auditory cues with linguistic and world knowledge to perform higher-level inference, remains an open question. We leave this in-depth exploration for future work.

\section{Acknowledgement}
We thank the reviewers for their constructive feedback during the review process, which greatly contributed to improving this work. We also acknowledge the computational and storage support provided by the National Center for High-performance Computing (NCHC) of the National Applied Research Laboratories (NARLabs) in Taiwan.

\bibliographystyle{IEEEtran}
\bibliography{IEEEexample}

\end{document}